\documentclass[dvipsnames,format=sigconf]{acmart}

\AtBeginDocument{%
  }

\setcopyright{acmlicensed}
\copyrightyear{2024}
\acmYear{2024}
\setcopyright{acmlicensed}\acmConference[GECCO '24 Companion]{Genetic and Evolutionary Computation Conference}{July 14--18, 2024}{Melbourne, VIC, Australia}
\acmBooktitle{Genetic and Evolutionary Computation Conference (GECCO '24 Companion), July 14--18, 2024, Melbourne, VIC, Australia}
\acmDOI{10.1145/3638530.3664166}
\acmISBN{979-8-4007-0495-6/24/07}

\usepackage{newfloat}
\usepackage{caption}
\usepackage{multirow}
\usepackage{makecell}
\DeclareFloatingEnvironment[fileext=frm,placement={!ht},name=Frame]{myfloat}

\usepackage{lipsum}
\usepackage[framemethod=TikZ]{mdframed}
\usepackage{hyperref}

\newenvironment{frameenv}[1][]
{\def\mycaption{#1}\begin{myfloat}[tb]
        \begin{mdframed}[roundcorner=10pt,backgroundcolor=white]
        }
        {\end{mdframed}\expandafter\caption{\expandafter\mycaption}\end{myfloat}
}

\usepackage{amsmath}
\begin{document}

\title[Towards Evolutionary-based AutoML for Small Molecule PK Prediction]{Towards Evolutionary-based Automated Machine Learning for Small Molecule Pharmacokinetic Prediction}

\author{Alex G. C. de S\'{a}}
\affiliation{%
  \institution{Baker Heart and Diabetes Institute}
  \city{Melbourne}
  \state{Victoria}
  \country{Australia}
  \postcode{3004}
}
\affiliation{%
  \institution{\vspace{0.10cm}School of Chemistry \& Molecular Biosciences, \\The University of Queensland}
  \city{Brisbane City}
  \state{Queensland}
  \country{Australia}
  \postcode{4067}
}
\affiliation{%
  \institution{\vspace{0.10cm}The University of Melbourne}  
  \city{Parkville}
  \state{Victoria}
  \country{Australia}
  \postcode{3010}
}
\email{Alex.deSa@baker.edu.au}


\author{David B. Ascher}
\affiliation{%
  \institution{Baker Heart and Diabetes Institute}
  \city{Melbourne}
  \state{Victoria}
  \country{Australia}
  \postcode{3004}
}
\affiliation{%
  \institution{\vspace{0.1cm}School of Chemistry \& Molecular Biosciences, \\The University of Queensland}
  \city{Brisbane City}
  \state{Queensland}
  \country{Australia}
  \postcode{4067}
}
\affiliation{%
  \institution{\vspace{0.1cm}The University of Melbourne}  
  \city{Parkville}
  \state{Victoria}
  \country{Australia}
  \postcode{3010}
}
\email{d.ascher@uq.edu.au}

\renewcommand{\shortauthors}{de S\'{a} et al.}

\begin{abstract}
Machine learning (ML) is revolutionising drug discovery by expediting the prediction of small molecule properties essential for developing new drugs. These properties –  including absorption, distribution, metabolism and excretion (ADME) –  are crucial in the early stages of drug development since they provide an understanding of the course of the drug in the organism, i.e., the drug's pharmacokinetics. However, existing methods lack personalisation and rely on manually crafted ML algorithms or pipelines, which can introduce inefficiencies and biases into the process. To address these challenges, we propose a novel evolutionary-based automated ML method (AutoML) specifically designed for predicting small molecule properties, with a particular focus on pharmacokinetics. Leveraging the advantages of grammar-based genetic programming, our AutoML method streamlines the process by automatically selecting algorithms and designing predictive pipelines tailored to the particular characteristics of input molecular data. Results demonstrate AutoML's effectiveness in selecting diverse ML algorithms, resulting in comparable or even improved predictive performances compared to conventional approaches. By offering personalised ML-driven pipelines, our method promises to enhance small molecule research in drug discovery, providing researchers with a valuable tool for accelerating the development of novel therapeutic drugs.

\end{abstract}

\ccsdesc[500]{Computing methodologies~Machine learning}
\ccsdesc[500]{Applied computing~Bioinformatics}
\ccsdesc[500]{Computing methodologies~Search methodologies}
\ccsdesc[500]{Theory of computation~Evolutionary algorithms}

\keywords{AutoML, Bio(chem)informatics, Grammar-based Genetic Programming, Small Molecules, Pharmacokinetics}

\maketitle

\section{Introduction}

Machine learning (ML) has been aiding, enhancing, and accelerating drug discovery for several years \cite{Mak2023, Serghini2023}, including the development of predictive methods for pharmacokinetics (PK) \cite{Jang2001, Yu2010, Ruiz2008, Hasselgren2024}, pharmacodynamics (PD) \cite{Yu2010, Hasselgren2024}, and pharmacogenomics~(PG)~\cite{Pirmohamed2023}. This aspect has leveraged ML-driven drug discovery, which targets the proposal of \emph{in silico} (or computational) predictive methods for prescreening and, consequently, prioritising a set of compounds~\cite{deSa2022, Pires2015, Xiong2021}. 

Small molecule research plays a significant role in \emph{in silico} drug development and discovery.\footnote{Small molecules are low molecular weight (900--1,000 \emph{Daltons}) organic compounds~\cite{Lenci2020}.} Small molecules account for over 90\% of currently marketed drugs \cite{Southey2023, Makurvet2021}. Given their importance and broad use, a special focus will be given to small molecules in this work. In particular, this work will be centred on assessing PK properties of small molecule drugs, given that these properties dictate the course of the compounds in the organism, relating how they affect or interact with organs, cells and tissues \cite{Jang2001, Yu2010, Ruiz2008, Hasselgren2024}.

To derive PK (and toxicity) understanding, \emph{in silico} methods – e.g., SwissADME \cite{Daina2017}, ADMETlab 1.0 and 2.0 \cite{Dong2018, Xiong2021}, pkCSM \cite{Pires2015}, toxCSM~\cite{deSa2022}, admetSAR 1.0 and 2.0~\cite{Cheng2012, Yang2019}, Interpretable-ADMET~\cite{Wei2022}, and Deep-PK~\cite{Myung2024} – concentrate in the construction of new ML models for discovering small molecules with proper pharmacokinetics (PK) (and also toxicity) properties. This understanding comes from the exploration and optimisation of absorption, distribution, metabolism, excretion and toxicity (ADMET) parameters for small molecular data \cite{Ferreira2019}. Therefore, specific ML models are designed to predict each ADMET parameter separately, e.g., employing a single-label classification approach, or several parameters simultaneously, e.g., using a multi-label classification approach \cite{Yang2019}. These ML models might be useful in drug discovery pipelines to filter out molecules with non-desirable ADMET parameters (e.g., low metabolism and excretion) while keeping top-level ADMET molecular data (e.g., those with high safety standards or proper absorption).

Although the field of small molecule research has progressed extensively, most ML-driven ADMET pipelines and methods are manually designed and do not have their algorithms, hyper-parameters, and derived ML models automatically selected and optimised. This design approach yields bias, inefficiency and lack of predictive generalisability depending on the type of molecular data. In addition, small molecule representation is not investigated carefully, even though it is a crucial step in generating good descriptive features for ML. Moreover, these previous methods are static, not allowing the personalisation of ML models for new input data, limiting their applicability to real-world PK data, where an enormous quantity of data is constantly generated by pharmaceutical companies.

Accordingly, this work introduces a novel automated machine learning (AutoML) method~\cite{Hutter2019} for small molecule pharmacokinetic prediction. In this AutoML method, we conceptualised a context-free grammar~\cite{Sipser2012} to describe the search space of algorithms associated with molecular representation~\cite{Wigh2022, Raghunathan2022, David2020}, and feature preprocessing, feature selection, machine learning modelling and hyper-parameter optimisation of the respective ML algorithms~\cite{Raschka2019}. Next, a grammar-based genetic programming (GGP) method~\cite{Mckay2010} is designed, developed and employed to search for the best predictive pipeline for the respective small molecule dataset, utilising the previously defined search space.
Given that, our proposed AutoML method provides a seamless end-to-end approach, where the molecular data is the target to recommend new optimised ML-driven drug discovery pipelines

Results for 12 PK datasets demonstrate that our proposed AutoML method is capable of selecting a diverse range of ML algorithms for pharmacokinetic prediction while resulting in comparable or improved predictive performances to non-sophisticated search methods and state-of-the-art methods. We believe that our proposed method will yield important and proper resource for compound screening tools for small molecule research on drug discovery.

\section{Related Work}

\begin{figure*}[!htbp]
  \centering
   \includegraphics[scale=0.40]{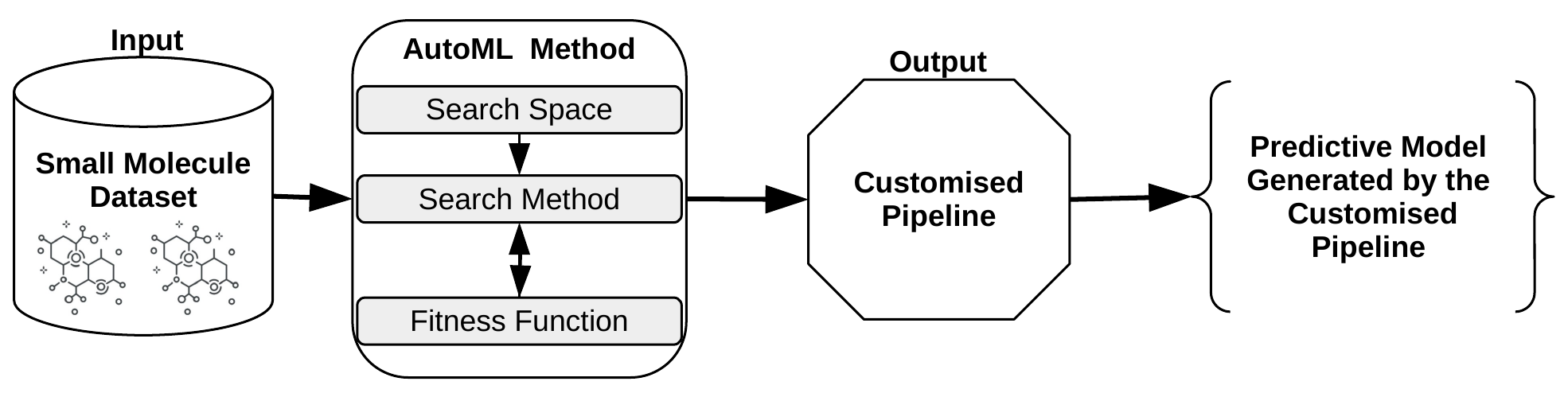}
   \caption{The general AutoML framework to select and configure ML pipelines for PK property prediction. Adapted from \cite{deSa2020}.}
   \label{Figure1}
 \end{figure*}

We divide this section into three main parts. First, we discuss related work that uses or applies evolutionary algorithms for small molecule pharmacokinetic (and toxic prediction) at some level. Next, we cover non-evolutionary state-of-the-art methods that apply machine learning to this task. Finally, we address a few approaches that target the proposal or evaluation of AutoML methods for predicting PK and toxicity properties of small molecules.

\subsection{Related work on Evolutionary Algorithms for Small Molecule PK and Toxicity Prediction }

Evolutionary algorithms are commonly used in PK and toxicity prediction to find the best set of features for a particular molecular dataset. For example, Soto et al. \cite{Soto2008a} proposed a genetic algorithm~(GA) \cite{Holland1992} for finding the best set of descriptors (i.e., features) in the context of ADMET prediction models, using decision trees~\cite{Breiman1984}, k nearest neighbours \cite{Aha1991} and polynomic non-linear function~\cite{Madsen2004} regression models to assess the quality of the feature set derived from a particular hydrophobicity molecular dataset \cite{Yaffe2002}. Soto et al. \cite{Soto2008b} extended their approach to the same problem, utilising a multi-objective version of the previously proposed GA in different work. More specifically, the authors used NSGA-II and SPEA2 Pareto-based strategies \cite{Deb2002, Zitzler2001} for finding the best feature set for a hydrophobicity molecular dataset.

Alternatively, Liu et al. \cite{Liu2018} introduced ECoFFeS, a method and software created for a user-friendly evolutionary-based feature selection for drug discovery. In Liu et al.'s work,  hERG blockers and logD7.4 molecular datasets were used to assess the applicability of ECCOFFeS. It is worth noting that ECOFFeS may be able to use single-objective evolutionary algorithms (SOEAs) or multi-objective evolutionary algorithms (MOEAs) for classification and regression problems. SOEAs in ECOFFeS include ant colony optimization (ACO)~\cite{Dorigo1996},  differential evolution (DE)~\cite{Storn1997},  genetic algorithm (GA) \cite{Holland1992} and particle swarm optimization (PSO) \cite{Kennedy1995}, while ECOFFeS incorporates two well-known MOEAs (i.e., MOEA/D \cite{Zhang2007} and NSGA-II~\cite{Deb2002}).

Evolutionary computation (EC) is also frequently employed to optimise small molecule structures, analyse molecule conformations, identify molecule superposition and detect their pharmacophores~\cite{Fromer2023, Lameijer2005}, for instance. In small molecule optimation, in particular, evolutionary operators can be applied in a straightforward manner. E.g., the mutation operator may be used to add or remove atoms, bonds or molecular fragments, while the crossover operator may be used to exchange molecular fragments among molecules \cite{Fromer2023}.

Other applications of EC in drug discovery, in general, can be found in the survey of Yu et al.~\cite{Yu2024}. This includes the use of EC for lead compound generation, such as ligands, and for exploring the quantitative structure-activity/property relationships of compounds.

\subsection{Related work on ML-Driven Small Molecule Research}

SwissADME \cite{Daina2017}, ADMETlab 1.0 and 2.0 \cite{Dong2018, Xiong2021}, pkCSM \cite{Pires2015}, admetSAR 1.0 and 2.0~\cite{Cheng2012, Yang2019}, toxCSM~\cite{deSa2022}, Interpretable-ADMET~\cite{Wei2022} and Deep-PK~\cite{Myung2024} stand as current and state-of-the-art ML methods for predicting pharmacokinetic (and/or toxicity) properties of small molecules. They primarily differ from each other in the number of molecular datasets, descriptors and machine learning methods.

SwissADME uses up to 23 molecular and physicochemical descriptors to build six traditional ML models targetting PK small molecule properties. pkCSM, in turn, employs the concepts of graph-based signatures, molecular descriptors and toxicophores to derive 31 PK and toxicity ML-based models for small molecules. Following the same principles of pkCSM, toxCSM was proposed to deal specifically with the toxicity properties of small molecules, incorporating 36 predictive models for this task. admetSAR 1.0, admetSAR 2.0 and ADMETlab 1.0 also used descriptors and traditional machine learning to construct 27, 47 and 31 PK and toxicity models, respectively. 

Applying deep learning models for small molecule PK and toxicity prediction is currently a hot area. In fact, ADMETlab 2.0, Interpretable-ADMET and Deep-PK derived 53, 59 and 73 robust deep learning models to predict the properties of small molecules, respectively. 

\subsection{Related work on AutoML for Small Molecule PK and Toxicity Prediction}

In spite of the common use of machine (and deep) learning to build models and methods for predicting a range of endpoint properties of small molecules, the automation of the model's choices and decisions, including the choice of the adequate molecular representation and features, is still an open problem. A few methods -- e.g. AutoQSAR \cite{Dixon2016},  ZairaChem \cite{Turon2023}, Uni-QSAR \cite{Gao2023} and Qptuna \cite{Mervin2024}-- have been used to investigate how optimisation and automation can be used to address these aspects. 

AutoQSAR employs an accuracy-based score to rank the ML pipelines that aim to solve the quantitative structure-activity relationship (QSAR) between compound and ADMET properties. However, AutoQSAR seems to apply an exhaustive search on the molecular datasets, making it not scalable depending on the number of ADMET tasks. ZairaChem, on the other hand, is an open-source AutoML package to derive a screening tool for drug candidates. ZairaChen uses five AutoML methods independently, including FLAML \cite{Wang2021}, aiming to specify different features (e.g., predictive performance, interpretability and robustness) to create 14 personalised ML models targetting molecule's PK, toxicity and other properties. Finally, Uni-QSAR and Qptuna are automated QSAR frameworks for molecule property prediction. While Uni-QSAR uses stacking to ensemble ML models and predicts molecular properties, Qptuna employs Bayesian optimisation for the same task.

To the best of our knowledge, no previous work has exploited the characteristics of evolutionary algorithms in AutoML to properly predict the PK properties of small molecules. Other evolutionary-based methods -- e.g.,  RECIPE (REsilient ClassifIcation Pipeline Evolution) \cite{deSa2017}, Auto-MEKA \cite{deSa2020} and TPOT (Tree-Based Pipeline Optimization Tool) \cite{Olson2016}  -- have been proposed for general AutoML. However, they have not been designed to take into account the characteristics of small molecule data and tasks. In the next section, we present our proposed approach that leverages the characteristics of a grammar-based genetic programming method to search and optimise ML pipelines in the context of small molecule research.

\section{AutoML Method for Predicting Small Molecule Pharmacokinetic Properties} \label{automl}

This section introduces the proposed AutoML method developed to search and optimise ML pipelines in the context of small molecule PK prediction. Section \ref{general_workflow} describes the main workflow and components followed by the method. Next, we present the three main aspects of the AutoML method: the \emph{search space}, the \emph{search method} and the \emph{fitness function}. 

In Section \ref{search_space}, we introduce the designed \emph{search space}, which covers the algorithmic choices for providing the chemical representations, ML algorithms and hyper-parameters. Moreover, Section  \ref{search_method} highlights the main aspects of the developed \emph{search method}, which is employed to search over the \emph{search space}. Finally, Section \ref{fitness_function} characterises the \emph{fitness function} to quantify how good an ML pipeline in the \emph{search space} is to guide the exploration and exploitation of the \emph{search method}.

\subsection{General Workflow} \label{general_workflow}

As depicted in Figure \ref{Figure1},  the proposed AutoML method receives as input a specific small molecule dataset with the pharmacokinetic~(PK) experimental targets. As aforementioned, the proposed AutoML methods have three main components: \emph{search space}, \emph{search method} and a \emph{fitness function}. The designed \emph{search space} covers the main algorithmic components (e.g., molecular representation techniques, threshold values, hyper-parameters, scaling and feature selection methods and the ML algorithms) from previously designed ML pipelines for PK prediction. To explore this \emph{search space}, the AutoML method takes advantage of a \emph{search method} and a \emph{fitness function}. The \emph{search method} employs a \emph{fitness function} to weigh the quality of each searched pipeline in the explored \emph{search space} and discover appropriate ML pipelines for the small molecule dataset at hand. Nevertheless, the predictive performance of the \emph{search method} naturally relies on what is specified in the \emph{search space} and how the \emph{fitness} function weighs the pipelines. 

The AutoML method outputs an ML pipeline tailored to the input dataset based on that \emph{search space}. This ML pipeline is specifically selected and (hyper-)parameterised to those input small molecule data, although it could be applied to other small molecule datasets. Finally, the customised pipeline returns a predictive model and, consequently, its classification (or regression) results.

The proposed AutoML method for small molecule PK prediction is freely available at \url{https://github.com/alexgcsa/ecada_2024/}.

\subsection{Search Space} \label{search_space}

  \begin{frameenv}[The excerpt of the proposed AutoML grammar.\label{frame1}]
       <Start> ::= <feature\_definition> [<feature\_scaling>] [<feature\_selection>] <ML\_algorithms> \vspace{0.1cm}
       
       <feature\_definition> ::= General\_Descriptors | Advanced\_Descriptors |
                       Graph-based\_Signatures | Toxicophores |
                       Fragments | General\_Descriptors Advanced\_Descriptors | General\_Descriptors Graph-based\_Signatures | 
                       ... | General\_Descriptors Advanced\_Descriptors
                       Graph-based\_Signatures Toxicophores 
                       Fragments \vspace{0.3cm}
                       
        <feature\_scaling> ::= <Normalizer> | <MinMaxScaler> | <MaxAbsScaler> | <RobustScaler> | <StandardScaler> \vspace{0.1cm}

        <Normalizer> ::= Normalizer <norm>
        
        <norm> ::= l1 | l2 | max\vspace{0.1cm}
        
        ... \vspace{0.3cm}

        <StandardScaler> ::= StddScaler <with\_mean> <with\_std>

        <with\_mean> ::= True | False
        
        <with\_std> ::= True | False \vspace{0.3cm}

        <feature\_selection> ::= <Variance\_Threshold> | <Select\_Percentile> | <SelectFPR> | <SelectFWE> | <SelectFDR> \vspace{0.1cm}
        
        <Variance\_Threshold> ::= VarianceThreshold <threshold>

        <threshold> ::= 0.0 | 0.05 | 0.10 | 0.15 | ... | 0.85 | 0.90 | 0.95 | 1.0\vspace{0.1cm}
        
        ... \vspace{0.3cm}
        
        <ML\_algorithms> ::= <AdaBoost> | <DecisionTree> | <ExtraTree> | <RandomForest> | <ExtraTrees> | <XGBoost>\vspace{0.1cm}

        <AdaBoost> ::= AdaBoost <algorithm> <n\_estimators> <learning\_rate>

        <algorithm> ::= SAMME.R | SAMME
        
        <n\_estimators> ::= 5 | 10 | 15 | 20 | ...| 300 | 500 | 550 | ... |  950 | 1000 | 1500 | 2000 | 2500 | 3000
        
        <learning\_rate> ::= 0.01 | 0.02 | 0.03 | ... | 2.0\vspace{0.1cm}
        
        ... \vspace{0.3cm}

        <XGBoost> ::= XGBoost <n\_estimators> <max\_depth> <max\_leaves> <learning\_rate>
        
        <max\_depth> ::= 1 | 2 | 3 | 4 | 5 | 6 | 7 | 8 | 9 | 10 | None
        
        <max\_leaves> = 1 | 2 | 3 | 4 | 5 | 6 | 7 | 8 | 9 | 10
    \end{frameenv}

The designed \emph{search space} is excerpted in Frame \ref{frame1}.\footnote{The complete grammar is available at \url{https://github.com/alexgcsa/ecada_2024/tree/main/bnf/grammar.bnf}.} This \emph{search space} is defined by a context-free grammar, which considers four-tuple <N, T, P, S>.  In this grammar, N is a set of non-terminals, T is a set of terminals, P is a set of production rules, and S (a member of N) is the start symbol. The set of production rules derives the language by combining the grammar symbols. In the grammar, the symbol "|" represents a choice, and the non-terminals surrounded by the symbols "[" and "]" are optional, i.e., they can appear or not in the production rules.

The start rule -- <Start> in the grammar in Frame \ref{frame1} -- defines the four main components of the PK prediction pipeline: (i) the molecular representation (defined by the non-terminal <feature\_definition>), (ii) the feature scaling, (iii) the feature selection, and (iv) the machine learning modelling (defined by the non-terminal <ML\_algorithms).

In molecular representation, 31 combinations of small molecule representation techniques are available. The main groups are molecular descriptors, advanced molecular descriptors, graph-based signatures, fragments, and toxicophores \cite{Pires2015, deSa2022}. This \emph{search space} component basically defines the features that are used to represent the small molecules, based on their biochemical structure.

Next, in feature scaling, typical scalers from the scikit-learn library were employed \cite{Pedregosa2011, Raschka2019}, including Normalizer, Min Max Scaler, Max Abs Scaler, Robust Scaler and Standard Scaler. This component basically modifies the variables representing the small molecules, setting distinct ranges for their values. However, there is also the option of not using any scaling technique on the features representing the small molecule data, which can be found in the <Start> rule of the grammar.

The other component involves selecting the produced features with a feature selection method \cite{Raschka2019}. The grammar is composed by the following methods from scikit-learn: Variance Threshold, Select Percentile, Select based on a False Discovery Rate (FDR), Select based on a False Positive Rate (FPR), and Select based on a  Family-Wise Error (FWE) rate. Similar to scaling, there is an option of not using any feature selection method, which is defined in the <Start> grammar rule.

In the currently proposed grammar, only classification pipelines are considered as part of grammar modelling \cite{Raschka2019}. At the moment, the ML modelling includes six algorithms from scikit-learn and independent software \cite{Pedregosa2011, Chen2016}: Decision Tree, Extremely Randomised Tree (Extra Tree), Random Forest, Extremely Randomised Tree (Extra Trees), Adaptive Boosting (AdaBoost), and Extreme Gradient Boosting (XGBOOST).

Taking into account these options and their respective hyper-parameters, the designed AutoML grammar has a total of 25 (non-redundant) production rules, with 24 non-terminals and 317 terminals.

\begin{figure*}[!htbp]
  \centering
   \includegraphics[scale=0.35]{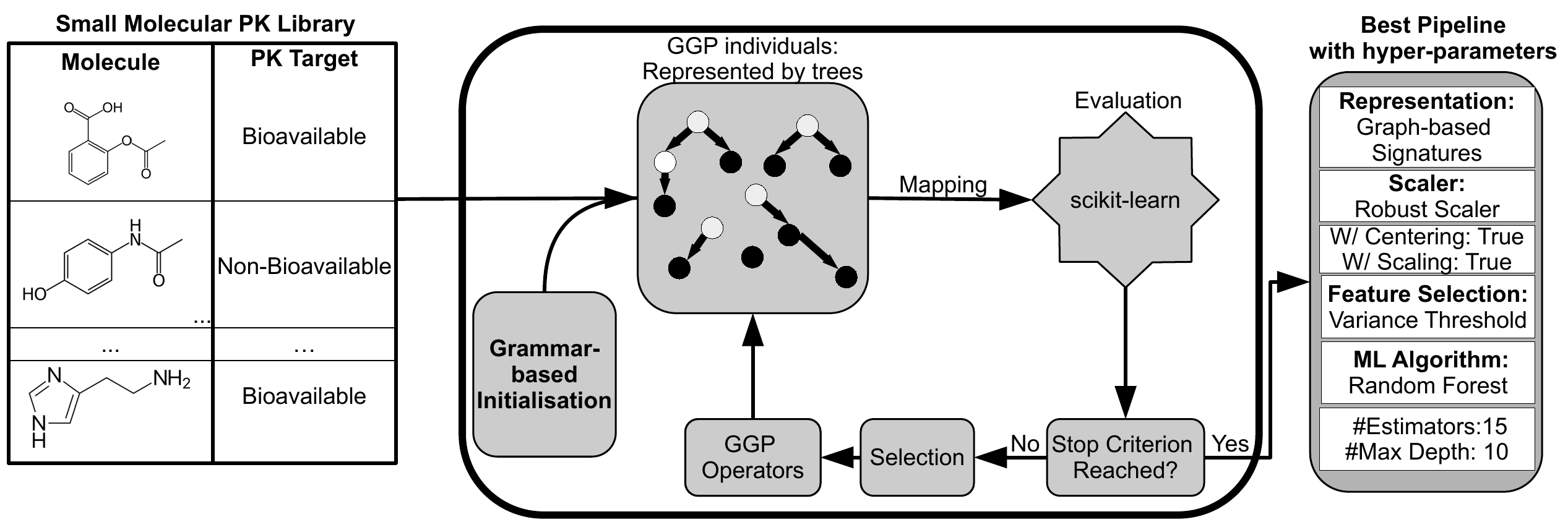}
   \caption{The grammar-based genetic programming (GGP) method to search for ML pipelines in the context of PK property prediction. Adapted from \cite{deSa2017}.}
   \label{fig-general}
 \end{figure*}

\subsection{Search Method}  \label{search_method}

A grammar-based genetic programming (GGP) method was defined to search and configure ML pipelines for predicting the pharmacokinetic properties of small molecules. As illustrated in Figure \ref{fig-general}, this method starts its evolutionary process by receiving a small molecule dataset as its input. This dataset contains the set of molecules and their respective experimental assay targets (e.g., the bioavailability of that respective molecule \cite{Daina2017,Dong2018, Xiong2021,Pires2015,Cheng2012, Yang2019,Wei2022,Myung2024}).

In the next step, the GGP utilises the grammar to initialise the first population at random. Given the simple initialisation, the individuals (i.e., the pipelines) are represented by parse trees. Thereafter, they are transformed into a string array, where they are subsequently mapped into a pipeline. Part of these pipelines are related to finding good representations of the small molecules. The other part is assigned to data science or machine learning methods (e.g., feature scaling, feature selection and machine learning components). After correctly mapping these pipelines, they are evaluated using the scikit-learning library.

This evaluation sets the pipelines' fitness scores. The following step includes checking if the stopping criterion is reached. If the stopping criterion is not reached, the pipelines pass through selection based on the fittest individuals and GGP operators (i.e., Whigham's crossover and mutation \cite{Mckay2010}). Tournament selection is applied to select the individuals, where rewards based on higher fitness scores are given to individuals during the selection process. After that, the GGP operators are applied one after the other. Therefore, if both operators occur, the mutation acts on the recombined pipelines derived from the crossover operation. It is also worth noting that the GGP operators, crossover and mutation, only produce valid individuals in accordance with the grammar. In addition, elitism is employed where the $n$ best pipelines from the previous generation are kept. 
This evolutionary process, shown in Figure \ref{fig-general}, is repeated until the stopping criterion is met. When this occurs, the best pipeline based on the last evaluated population is returned together with the most suitable hyper-parameters, which were also found by the GGP method.

\subsection{Fitness Function}  \label{fitness_function}

\begin{figure}[!htbp]
  \centering
   \includegraphics[scale=0.34]{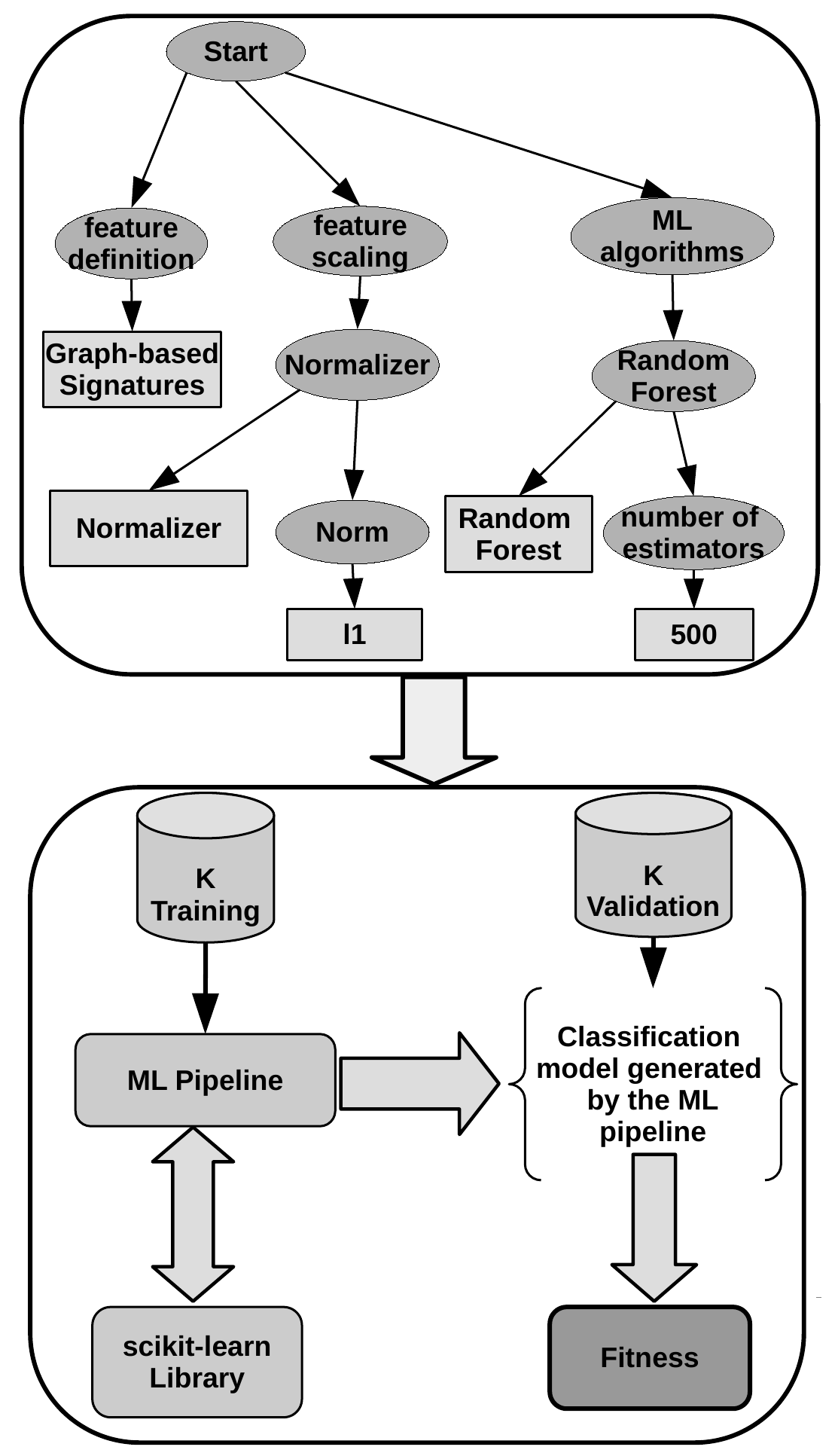}
   \caption{The evaluation process in the GGP-based AutoML method to assess ML-driven PK pipelines. Adapted from \cite{deSa2020}.}
   \label{fig-eval}
 \end{figure}

Figure \ref{fig-eval} encompasses an example of the evaluation process to provide a fitness score for each searched pipeline (i.e., the individuals of the GGP method) in more detail. In Figure \ref{fig-eval}, ellipsoids map non-terminals, while rectangles determine terminals. Figure \ref{fig-eval} basically depicts how the parse trees representing the individuals are mapped into an ML pipeline. 

In the first derivation from the \emph{Start} node -- i.e., in \emph{feature definition} -- molecular data is transformed into tabular representations, where the table's columns denote features. In Figure \ref{fig-eval}, features assume only descriptors, called Graph-based Signatures, from graphs representing the molecules.

Next, these generated features can be scaled and/or selected. In the example of Figure \ref{fig-eval}, features are only scaled (i.e., based on the \emph{feature scaling node}) with the method Normalizer from scikit-learn, where a norm (e.g., l1) needs to be fixed. Given the dataset is ready, we go for ML modelling using \emph{ML Algs}. In this part, a machine learning algorithm (e.g., Random Forest with 500 random decision tree estimators) is used to build the predictive model for that data. 

In the conceptualised evaluation process, we employ a K-fold cross-validation procedure to estimate the quality of the pipeline~\cite{Raschka2018}. Therefore, the pipeline is run K times, changing the training and validation sets in each iteration. Considering the training and validation sets, the scikit-learn library assesses the model's performance. This results in a classification model, where the fitness is extracted as an average from its predictions on the K validation sets.

For estimating the quality of each pipeline, the \emph{fitness fuction} is set as the average of Matthew’s correlation coefficient (MCC)~\cite{Chicco2020} over a 5-fold cross-validation. MCC is a performance measure, which is used in classification tasks. It is especially relevant and robust in cases where data imbalance is present. MCC is defined in Equation \ref{MCC}.

\begin{equation} \label{MCC}
    MCC = \frac{((TP \times TN) - (FP \times FN))}{\sqrt{(TP + FP) \times (TP + FN) \times (TN + FP) \times (TN + FN)}}
\end{equation}

In Equation \ref{MCC}, TP represents the true positives (molecules labelled as 1, predicted as 1), TN the true negatives (molecules labelled as 0, predicted as 0), FP the false positives (molecules labelled as 0, predicted as 1) and FN the false negatives (molecules labelled as 0, predicted as 1).

MCC considers all sample types (i.e., TP, TN, FP and FN) to result in a measure that ensures a fair evaluation of imbalance learning. As MCC is a correlation coefficient, it can vary from -1.0 to +1.0. +1.0 denotes a perfect positive correlation, and -1.0 represents a perfect inverse correlation. In addition, a classifier is considered random if its MCC is equal to 0.0.

\section{Experiments}

This section conceives the main aspects of the AutoML experiments for pharmacokinetic small molecular data, including the description of the datasets (Section \ref{datasets}), the configurations of the grammar-based genetic programming (GGP) method (Section \ref{configuration}), and the comparison to alternative methods (Section \ref{comparisons}).

\subsection{Datasets}\label{datasets}

12 PK datasets were used to estimate the quality of the AutoML method. They encompass binary classification tasks associated with absorption, metabolism and excretion experimental \emph{in vivo} or \emph{in vitro} tests on small molecules. These PK datasets vary in terms of their respective number of small molecules (\# Molecules), ranging from 404 to 18,558. This variation in terms of the size of the dataset imposes a challenge to the AutoML method to handle the searched pipelines.

\begin{table}[htbp]
    \scriptsize
    \centering
    \caption{Description of 12 binary classification PK datasets.}
    \begin{tabular}{|l|p{2.5cm}|l|l|l|}
        \hline
        \textbf{ID} & \textbf{Dataset} & \textbf{Abbreviation} & \textbf{Category} & \textbf{\# Molecules} \\
        \hline
        1 & Caco-2 permeability & Caco-2 & Absorption & 663 \\ \hline
        2 & P-glycoprotein I Inhibitor & PGP I Inhibitor & Absorption & 1223 \\ \hline
        3 & P-glycoprotein II Inhibitor & PGP II Inhibitor & Absorption & 1023 \\ \hline
        4 & P-glycoprotein I Substrate & PGP I Substrate & Absorption & 1272 \\ \hline
        5 & Skin Permeability & Skin Perm. & Absorption & 404 \\ \hline
        6 & Cytochrome P450 CYP2C9 Inhibitor & CYP2C0 Inhibitor & Metabolism & 14,706 \\ \hline
        7 & Cytochrome P450 CYP2C19 Inhibitor & CYP2C19 Inhibitor & Metabolism & 14,572 \\ \hline
        8 & Cytochrome P450 CYP2D6 Inhibitor & CYP2D6 Inhibitor & Metabolism & 14,738 \\ \hline
        9 & Cytochrome P450 CYP2D6 Substrate & CYP2D6 Substrate & Metabolism & 666 \\ \hline
        10 & Cytochrome P450 CYP3A4 Inhibitor & CYP3A4 Inhibitor & Metabolism & 18,558 \\ \hline
        11 & Cytochrome P450 CYP3A4 Substrate & CYP3A4 Substrate & Metabolism & 669 \\ \hline
        12 & Renal Organic Cation Transporter 2 Substrate & OCT2 Substrate & Excretion & 904 \\
        \hline
    \end{tabular}
    \label{tab:datasets}
\end{table}

We split the complete datasets into two sets (i.e., training data and blind test data) in a stratified way. 90\% of the dataset is used to search for the best pipeline for the respective PK dataset, whereas the remaining 10\% is set to assess the final selected pipeline's accuracy.

\subsection{GGP Parameter Configuration}\label{configuration}

The GGP parameters were configured in the following way. 100 individuals representing ML pipelines are evolved for one hour or at most 50 generations. Each individual has at most 5 minutes to run. Otherwise, its run is interrupted and its score is set to 0.0. Crossover and mutation operators are employed with a probability rate of 0.90 and 0.10, respectively. Over the generations, the best current individual is kept for the next generations (i.e., the elitism size is equal to 1). Finally, to avoid overfitting happening on the final model generated by the best pipeline, data going for cross-validation is resampled every five (5) generations. This way, the generated model would be more aware of data variations.

\begin{table}[htbp]

    \centering
    \caption{The GGP parameters for evolving a population of machine learning pipelines for PK prediction.}
    \begin{tabular}{|l|l|}
        \hline
        \textbf{Parameter} & \textbf{Value} \\
        \hline 
        Population Size & 100 \\ \hline
        Stopping Criterion & 1 hour \\ \hline
        Crossover Probability & 0.90 \\ \hline
        Mutation Probability & 0.10 \\     \hline    
        Elitism Size & 1 \\ \hline
        Data Resampling & Every 5 Generations \\ \hline
        Individual's time budget & 5 minutes \\
        \hline
    \end{tabular}
    \label{tab:configurations}
\end{table}

\subsection{Comparison} \label{comparisons}

We compared the best pipelines found by the method introduced in Section \ref{automl} with two alternative approaches. First, after running the proposed AutoML method 20 times -- in order to provide reliable statistical analysis -- and gathering the final pipelines, we compare their performance and the best AutoML-found pipeline on the 5-fold cross-validation procedure against pkCSM \cite{Pires2015}, which is a well-known method for predicting the PK properties of small molecules. Second, we contrast the selected pipelines and the best pipeline found by the proposed AutoML method against the XGBOOST \cite{Chen2016} using its default parameters. We decided on XGBOOST in the comparisons given its popularity in machine learning and its common use on a daily basis.

Finally, the Iman Davenport's modification of Friedman's test~\cite{Demvsar2006} is used to compare the four methods to each other. If this test is significant, it is followed by a Nemeyi \emph{post hoc} test, which is used to compare the predictive performances of the AutoML method and best-found pipeline with the alternative approaches statistically in a pairwise manner.

\section{Results}

This section presents the results considering the experiments detailed in the previous section. We first provide a summary of the predictive performance in Section \ref{summary}, supplying some aspects of the performance behaviour of the designed AutoML method. Following, we compare the proposed AutoML method against the baselines in Section \ref{comparison}. Finally, in Section \ref{analysis}, we analyse the best-selected algorithms across all datasets to hint at what is characterised as a good pipeline by the evolutionary search of our AutoML method for PK prediction.

\subsection{Summary of AutoML Predictive Performance} \label{summary}

Table \ref{tab:results} outlines the average of the predictive results of the selected pipelines in terms of MCC, followed by the standard deviation between parenthesis. Results in this table are divided between a 5-fold cross-validation (5-fold CV) and a blind test set for each dataset.

Overall, the results in Table \ref{tab:results} highlight a general generalisation trend for the AutoML method, meaning that in most cases the predictive performance to choose the ML pipeline (i.e., in the 5-fold CV) corresponds to the final predictive performance  (i.e., in the blind test set). This generalisation aspect emphasises the AutoML search and optimisation derived by the GGP method works consistently well.

There are a few cases where the results follow possible cases of overfitting, such as \emph{PGP II Substrate} and \emph{CYP2D6 Substrate}, even if we employed strategies to avoid it from occurring. These results might indicate a possible bias in the splitting of the datasets and that the cross-validation on the AutoML method is not assessing correctly the performance of the pipeline in some cases, needing further attention and investigation.

\emph{PGP II Inhibitor} and \emph{CYP3A4 Substrate}, on the other hand, present a different trend, where the results in cross-validation over the training set had lower MCC scores when compared to the blind test set. One possible explanation for this pattern is the fact that the data is resampled every 5 generations. Although the 5-fold CV results are lower, they prepared the pipeline for data variation, which happened in the blind test set.

\begin{table}[htbp]
    \centering
    \caption{The 5-fold cross-validation and blind test results of the proposed AutoML method in terms of MCC for the best-selected ML pipelines for the 12 PK datasets.}
    \begin{tabular}{|c|l|c|c|}
        \hline
        \textbf{ID} & \textbf{Dataset} & \textbf{5-fold CV} & \textbf{Blind Test} \\
        \hline
        1 & \makecell{Caco-2} & 0.589 (0.024) & 0.570 (0.042) \\  \hline
        2 & \makecell{PGP I \\ Inhibitor} & 0.786 (0.023) & 0.792 (0.044) \\ \hline
        3 & \makecell{PGP II \\ Inhibitor} & 0.617 (0.019) & 0.754 (0.037) \\ \hline
        4 & \makecell{PGP II \\ Substrate} & 0.494 (0.037) & 0.287 (0.107) \\ \hline
        5 & \makecell{Skin \\ Perm.} & 0.406 (0.025) & 0.420 (0.110) \\ \hline
        6 & \makecell{CYP2C9 \\ Inhibitor} & 0.565 (0.026) & 0.578 (0.028) \\ \hline
        7 & \makecell{CYP2C19 \\ Inhibitor} & 0.599 (0.028) & 0.619 (0.019) \\ \hline
        8 & \makecell{CYP2D6 \\ Inhibitor} & 0.506 (0.034) & 0.528 (0.024) \\ \hline
        9 & \makecell{CYP2D6 \\ Substrate} & 0.469 (0.018) & 0.284 (0.088) \\ \hline
        10 & \makecell{CYP3A4 \\ Inhibitor} & 0.528 (0.024) & 0.563 (0.024) \\ \hline
        11 & \makecell{CYP3A4 \\ Substrate} & 0.255 (0.013) & 0.427 (0.042) \\ \hline
        12 & \makecell{OCT2 \\ Substrate} & 0.475 (0.019) & 0.371 (0.062) \\
        \hline
    \end{tabular}
    \label{tab:results}
\end{table}

\subsection{Comparison against Alternative Methods} \label{comparison}

When comparing the proposed AutoML method against pkCSM and XGBOOST, the selected pipelines were able to reach consistently better average predictive scores in terms of MCC in the majority of the cases. This can be seen in Table \ref{tab:comparison}. This performance demonstrates that AutoML is able to select pipelines to generate models for pharmacokinetic prediction properly. 

It is worth noting that the AutoML method performed better than or equal to pkCSM, a well-known method for predicting PK, in 9 out of the 12 datasets. Similarly, the proposed AutoML method is better than or equal to XGBOOST in 8 out of the 12 Pk datasets. However, recall that the AutoML method's scores are an average of 20 repetitions, meaning we can find better pipelines for specific cases.

For this reason, we compare the best-selected pipeline by our novel AutoML method -- based on 5-fold cross-validation across its 20 runs -- to compare against pKCSM and XGBOOST. It is clear in Table \ref{tab:comparison} that the results improve predictive performance when using the best-selected AutoML pipeline. When specifically checking the reached MCC values, the best-selected pipelines perform better than or equal to pKCSM and XGBOOST in 10 and 11 out of the 12 cases, respectively.


Although this emphasises that AutoML can reach better predictive scores when selecting proper representations, scalers, feature selectors, and ML algorithms, it also shows that possible improvements can be incorporated into the search method to diminish overfitting and improve predictive performance.

\begin{table}[htbp]
    \centering
    \caption{Comparison of the proposed AutoML method and its best-selected pipeline across the 20 runs against pkCSM and XGBOOST (XGB) in terms of MCC metric on the blind test set.}
    \begin{tabular}{|c|l|c|c|c|c|}
        \hline
        \textbf{ID} & \textbf{\makecell{Dataset}} & \textbf{\makecell{Proposed \\ AutoML \\ Method}} & \textbf{\makecell{Best \\ AutoML- \\ Selected}} & \textbf{pkCSM} & \textbf{XGB} \\
        \hline
        1 & \makecell{Caco-2} & 0.570 & 0.610 & 0.609 & 0.579 \\  \hline
        2 & \makecell{PGP I \\ Inhibitor} & 0.792 & 0.837 & 0.776 & 0.820 \\ \hline
        3 & \makecell{PGP II \\ Inhibitor} & 0.754 & 0.783 & 0.716 & 0.696 \\ \hline
        4 & \makecell{PGP II \\ Substrate} & 0.287 & 0.289 & 0.214 & 0.232 \\ \hline
        5 & \makecell{Skin \\ Perm.} & 0.420 & 0.394 & 0.108 & 0.368 \\ \hline
        6 & \makecell{CYP2C9 \\ Inhibitor} & 0.578 & 0.615 & 0.601 & 0.553 \\ \hline
        7 & \makecell{CYP2C19 \\ Inhibitor} & 0.619 & 0.647 & 0.583 & 0.590 \\ \hline
        8 & \makecell{CYP2D6 \\ Inhibitor} & 0.528 & 0.556 & 0.408 & 0.488 \\ \hline
        9 & \makecell{CYP2D6 \\ Substrate} & 0.284 & 0.334 & 0.197 & 0.267 \\ \hline
        10 & \makecell{CYP3A4 \\ Inhibitor} & 0.563 & 0.590 & 0.623 & 0.534 \\ \hline
        11 & \makecell{CYP3A4 \\ Substrate} & 0.427 & 0.274 & 0.289 & 0.440 \\ \hline
        12 & \makecell{OCT2 \\ Substrate} & 0.371 & 0.427 & 0.353 & 0.402 \\
        \hline
        \multicolumn{2}{|c|}{\textbf{Average}} & 0.516 & 0.530 & 0.456 & 0.497 \\ \hline
        \multicolumn{2}{|c|}{\textbf{Ranking}} & 2.417 & 1.417 & 3.250 & 2.917 \\
        \hline
    \end{tabular}
    \label{tab:comparison}
\end{table}

Considering the modified Friedman's statistical test to compare all the methods against each other, it resulted in a significant p-value ($= 0.001041$), demonstrating that there are differences across the methods predicting PK properties. By running  Nemeyi \emph{post hoc} test based on the differences among the rankings (see Table \ref{tab:comparison}) compared to the found Nemeyi's critical difference ($= 1.4072$), we discovered that the AutoML method has no statistical difference to both pKCSM and XGBOOST. However, when comparing the best AutoML-selected pipeline to pkCSM and XGBOOST, a statistical difference is detected. This means the AutoML method is able to find pipelines that present statistically different performance to commonly used approaches, such as pkCSM or XGBOOST.

\subsection{Analysis of the Selected ML Pipelines} \label{analysis}

The predictive performance derived from the previous sections is brought by the selection of pipelines encompassing representation, feature scalers, feature selectors, and ML algorithms. Here, we analyse the most selected components for each main block from the grammar to see what is yielding performance on the proposed AutoML method.

Starting with representation, on average, the most selected representation is the combination of General Descriptors, Advanced Descriptors, and Graph-based Signatures (13\%), followed by the combination of Advanced Descriptors, Graph-based Signatures, and Fragments (9.6\%) and the combination of General Descriptors, Graph-based Signatures, Toxicophores, and Fragments (9.2\%). This result highlights that combining small molecule descriptors is advantageous when predicting the molecules' pharmacokinetics. However, some combinations end up being more relevant to certain types of properties than others, making their correct selection crucial. This justifies the use of AutoML to find the best small molecule representation techniques.

When taking a look at the selected scalers, we notice that the majority of the selected algorithms do not have a scaling method plugged into their pipeline. By analysing our \emph{search space}, we conclude that this might be because of the decision to use tree-based ML algorithms, which are robust to the data range. In future work, we intend to analyse the behaviour of the AutoML method when we extend the \emph{search space}, including algorithms that are sensible to the feature value variation.

The feature selection methods are extensively selected, but not using feature selection is a common choice by the AutoML method. In terms of the order of most selected methods, we have:
Select Fwe (22.5\%), Variance Threshold and No Feature Selection (17.9\%), Select FDR (15.8\%), Select FPR (13.8\%) and Select Percentile (12.1\%). In a nutshell, selecting the use of a feature selection might be very important depending on the type of features generated in the first part of the grammar (i.e., representation). Not using feature selection tends to happen when the dimensionality is low and there is no need to reduce the feature space, which happens in some scenarios of combinations of the feature tested. 

Finally, we analyse the selection of the ML algorithms across all 20 repetitions $\times$ 12 datasets. Gradient Boosting is the most selected algorithm (42.5\% of the cases), followed by Random Forest (20\%), Extremely Randomised Trees (17.5\%),  XGBOOST (15.4\%) and Adaptive Boosting (4.6\%). Simple and randomised decision tree models were never selected as the best-selected classification algorithm. Therefore, this means that AutoML prefers to select complex but highly predictive ML algorithms. This selection based only on performance is a component we can modify later on the \emph{search method}, incorporating a negative reward for pipeline complexity, for example.

\section{Conclusions and Future Work}

This paper conceptualises and introduces a novel AutoML method to handle the challenge of predicting the pharmacokinetic (PK) properties of small molecules. This AutoML method is centred on four main aspects of the ML-driven PK pipeline: molecular representation, scaling, feature selection and ML modelling. To derive a suitable AutoML method, we explored the use of a grammar to define a \emph{search space}, in which a grammar-based genetic programming method leveraged its search.

The results emphasise that AutoML can be used to find proper algorithms and building blocks to compose pipelines to predict PK properties. Our analyses show AutoML achieving better or comparable predictive results in terms of Matthew's correlation coefficient (MCC) against a state-of-the-art method, pkCSM, and a commonly used algorithm, XGBOOST. These comparison results are supported by a statistical test. Additionally, our attempt to understand the achieved performances also brought ideas of which possible algorithmic choices led to good predictive scores.

Although predictive results demonstrated a positive performance, we intend to further improve and explore the proposed AutoML method. The first modification we plan to study will be the extension of the current \emph{search space}, including new algorithms, methods and hyper-parameter choices. We also intend to redesign the evaluation process to consider certain characteristics of small molecules, such as cross-validation, taking into account non-redundancy and scaffolds (i.e., the core structures of small molecules). In addition, the AutoML method needs to be assessed in a broader range of small molecule datasets. This would secure a complete evaluation of the AutoML method. Finally, the GGP might enhance its predictive capabilities by using a surrogate to guide it through better parts of the search space. This is an option we target for future releases of the \emph{search method}.

\section*{Funding}
Investigator Grant from the National Health and Medical Research Council of Australia (GNT1174405); Victorian Government’s Operational Infrastructure Support Program (in part).

\bibliographystyle{ACM-Reference-Format}
\bibliography{abbrev.bib}

\end{document}